\documentclass[runningheads]{llncs}
\usepackage{graphicx}
\usepackage{listings}
\usepackage{pgfplots}
\usepackage{multicol}
\usepackage{smartdiagram}
\usepackage{metalogo}
\usepackage[edges]{forest}
\usepackage{tikz}
\usepackage{forest}
\pgfplotsset{compat=newest}
\usepackage{float}
\restylefloat{table}
\usepackage[colorinlistoftodos]{todonotes}
\usepackage{amsfonts}
\usepackage{booktabs}
\usepackage{siunitx}
\usepackage{colortbl} 
\usepackage{tabularx}

\begin{document}
\title{Machine Learning Meets Natural Language Processing - The story so far}
%
%
\author{Nikolaos-Ioannis Galanis\orcidID{0000-0001-9528-4349}\and
Panagiotis Vafiadis\orcidID{0000-0002-6448-5524}\and
Kostas-Gkouram Mirzaev\orcidID{0000-0003-4473-4631}\and
George A. Papakostas\orcidID{0000-0001-5545-1499}}

\authorrunning{Nikolaos-Ioannis Galanis et al.}
%
\institute{HUman-MAchines INteraction Laboratory (HUMAIN-Lab), Department of Computer Science, International Hellenic University, Kavala, Greece\\
\email{\{nigaean, pavazei, gkmerza, gpapak\}@cs.ihu.gr}}
%

\maketitle              
\begin{abstract}
Natural Language Processing (NLP) has evolved significantly over the last decade. This paper highlights the most important milestones of this period, while trying to pinpoint the contribution of each individual model and algorithm to the overall progress. Furthermore, it focuses on issues still remaining to be solved, emphasizing on the groundbreaking proposals of Transformers, BERT, and all the similar attention-based models.
\end{abstract}
\section{Introduction}

Records of NLP application can be found even before the early 1900's when there were attempts of using machine translation to translate text from one language to another \cite{brown2006the}. Meanwhile, there were some conflicting views between linguistics and computer science claiming that language is generative in nature and cannot be described with mathematical concepts \cite{pullum1945philosophy}.

Alan Turing adequately answered ``Can machines think?'' in 1950, by introducing the research/study of ``Imitation Game'' \cite{turing1950computing}, a simulation process of a computer acting and answering without substantially changing the outcome\cite{epstein2009parsing}. Thus the machine is considered to be ``thinking'', as long as having a conversation with it could be indistinguishable from that with a human. 

The first successful attempt to achieve that was ELIZA\cite{weizenbaum1966eliza}, a simple program within the Project of Mathematics and Computation (``Project MAC'') at MIT that managed to mislead people into believing that it's a psychologist, reflecting on questions by turning the questions back at the speaker. Another program was PARRY (Colby, 1975) mimicking a paranoid schizophrenic \cite{PinarSaygin2000}. Over the years, programs were getting "smarter" like Eugene Goostman \cite{goostman} or Cleverbot \cite{saenz_cleverbot_2010} that statistically analyzes huge databases of real conversations to determine the best responses.

The downside was the inability to keep consistency and keep up with brand new subjects. There are numerous variations or alternatives to the Turing test, like when humans have to prove their non-machine nature to a computer (ex. CAPTCHA \cite{von2003captcha}) or when we use AI to create original art.
\vspace{-1.5\baselineskip}
\begin{figure}[H]
\begin{frame}{}
\small
\begin{tikzpicture}[scale=0.25,every node/.style={outer sep=2pt}]
    \def\ourInfo{{
        {"1930","Translating machines patents, Artsourni, Troyanskii propose dictionaries"},
        {"1950","Imitation Game by Alan Turing"},
        {"1957","Syntactic Structures by Noam Chomsky"},
        {"1966","ELIZA: Computer Psychotherapist by Joseph Weizenbaum"},
        {"1968","SHRDLU: NLU Computer Program by Terry Winograd (MIT)"},
        {"1969","Conceputal Dependency theory by Roger Schank"},
        {"1975","PARRY: Computer Schizophrenic person by Colby "},
        {"1980","ATN: Augmented Transition Network by William A. Woods"}, 
        {"...", "MUBBLE,MOPTRANS,KODIAK,ABSITY,DR.SPAITSO,RACTER"},
        {"2006","AI Software (Question Answering system) from IBM  by Watson"},
        {"2011", "Siri : Mobile Assistant by Apple"},
        {"2013", "WORD2VEC by Tomas Mikolov et al"},
        {"2014", "GloVe by Jeffrey Pennington et al, Alexa : Assistant by Amazon"},
        {"2016", "FASTTEXT by Piotr Bojankwski et al"},
        {"2017", "Transformer by Vaswani et al, Chatbots used in business operations"},
        {"2018", "ELMO by Matthew E Peters et al, BERT by J. Devlin et al"},
    }}
    \pgfmathsetmacro{\length}{10}

    \foreach \i in {0, ..., \length}{
        \pgfmathsetmacro{\year}{\ourInfo[\i][0]}
        \pgfmathsetmacro{\eventName}{\ourInfo[\i][1]}
        \draw[thick,red] (0,-2*\i-2)--(0,-2*\i);
        \draw(0,-2*\i-1) node[black, right, align = left]{\eventName};
        \draw(0,-2*\i-1) node[black, left] {\year};
    }
    \foreach \i in {0, ..., \length}{
        \filldraw[draw = white, fill = red,thick] (0,-2*\i-1) circle (5pt);
        \draw[thick,red] (-12pt,-2*\i-1)--(0,-2*\i-1);
    }
\end{tikzpicture}
\end{frame}
\vspace{-1\baselineskip}
\caption{A summarized timeline of important NLP milestones.}
\end{figure}
\vspace{-1.5\baselineskip}
Over the last decade, NLP grew rapidly and led to next-gen applications, such as virtual assistants like Siri or Alexa. New methodologies were developed using neural networks or unsupervised learning for acquiring vector representations of words like Word2Vec or GloVe. The latest milestone in this growth, was the introduction of the attention-based models, using a mechanism that comprehends contextual associations between words and phrases.

NLP as a sub-field of AI, examines and detects patterns in data and uses them in achieving better understanding and generating natural language. There are several applications of NLP, some of them are:
\begin{multicols}{2}
\begin{enumerate}
    \item Search engines
    \item Virtual assistangs \& Chatbots
    \item Sentence segmentation
    \item Part of speech tagging
    \item Information Extraction
    \item Question Answering
    \item Machine Translation
    \item Deep Analysis
    \item Named entity recognition
    \item Spam Detection
    \item Text-to-speech \& Speech-to-Text
    \item Sentimental analysis
    \item Text Summarization
\end{enumerate}
\end{multicols}
NLP converts human language from the form of raw text data, into structured data (computer-understandable), but prior, it needs to perceive the data based on grammar, context and decide on intent and entities, with a process called Natural Language Understanding(NLU). On the other hand, Natural Language Generation(NLG), is a process that converts computer-generated data into human understandable text. This system generates well structured dynamic documents using both document-planning, micro-planning and realization, by representing human-like desired sentences\cite{manning2014natural}.


\section{Related Work}

Though other surveys have previously presented various trends in Natural Language Processing and Machine Learning (\cite{goldberg-review-2016},\cite{young-review-2018},\cite{otter-review-2021}), each in a different way either with a more practical or theoretical approach. In this review, we will attempt to pinpoint and summarize the most critical and important breakthroughs in the field of NLP up until today, while also focusing on the still existent and new emerging challenges.

\section{Materials and Methods}
\subsection{The Literature Accumulation}
The search begins with Google Scholar's highest cited results for the last 10 years, while also tracing references and backwards citations. The search included at least one keyword of each group: Machine Learning (Machine Learning, Transformer, CNN, Neural Network, Recurrent, GRU, Deep Learning, Recursive, LSTM, ML) and NLP (Natural Language, NLP, NLU, NLG).

Clearly, there's a steadily increasing number of papers on the combined subject, booming after the proposal of Transformers, with publications of derived models and methodologies for transfer learning (see Fig. \ref{fig:scholar_scopus_papers}).

\pgfplotsset{ every non boxed x axis/.append style={x axis line style=-},
     every non boxed y axis/.append style={y axis line style=-},
     yticklabel style={xshift=-1ex, anchor=east},
     legend columns=-1,
     axis y line*=left,
     axis x line*=bottom,
     xticklabel style={yshift=-1ex, anchor=east, rotate=40}}
     
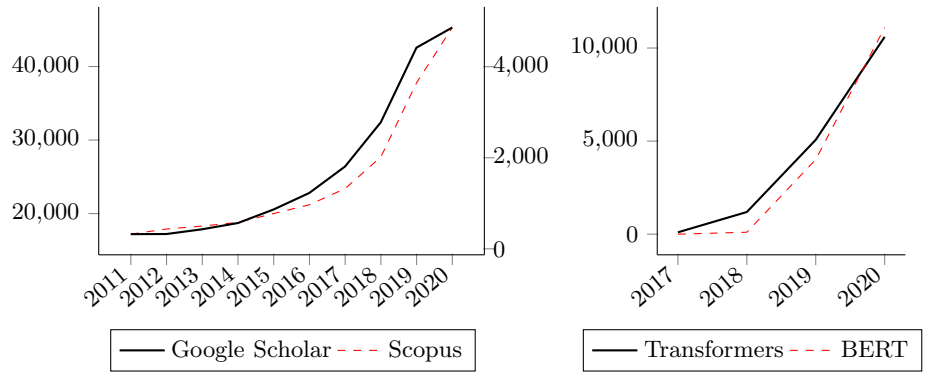
\begin{figure}[H]
    \begin{tikzpicture}
        \begin{axis}[
            scaled ticks=false,
            width = 0.55\linewidth,
            height = 0.40\linewidth,
            xticklabels from table={papers_history.dat}{Year},
            xtick style={yshift=-1ex, anchor=south},
            ytick style={xshift=-1ex, anchor=east},
            xtick = data]
            \addplot[red,dashed] table [y = GoogleScholar, x=X]{papers_history.dat};
            \label{google_scholar_plot}
        \end{axis}
        \begin{axis}[
            axis y line*=right,
            axis x line=none,
            scaled ticks=false,
            legend style={at={(0.03,-0.3)}, anchor=north west},
            width = 0.55\linewidth,
            height = 0.40\linewidth,
            yticklabel style={xshift=1ex, anchor=west},
            ytick style={xshift=1ex, anchor=west}]
            \addplot[black,thick] table [y=Scopus,x=X]{papers_history.dat};
            \addlegendimage{/pgfplots/refstyle=google_scholar_plot}
            \addlegendentry{Google Scholar}
            \addlegendentry{Scopus}
        \end{axis}
    \end{tikzpicture}
    \begin{tikzpicture}
        \begin{axis}[
            scaled ticks=false,
            width = 0.40\linewidth,
            height = 0.40\linewidth,
            xticklabels = {2017,2018,2019,2020},
            xtick style={yshift=-1ex, anchor=south},
            ytick style={xshift=-1ex, anchor=east},
            legend style={at={(-0.3,-0.3)}, anchor=north west},
            xtick = data]
            \addplot[black,thick] coordinates{(2017,99) (2018,1190) (2019,5070) (2020,10600)};
            \addplot[red,dashed] coordinates{(2017,0) (2018,103) (2019,4010) (2020,11100)};
            \label{bert_transformer_plot}
            \addlegendimage{/pgfplots/refstyle=bert_transformer_plot}
            \addlegendentry{Transformers}
            \addlegendentry{BERT}
        \end{axis}
    \end{tikzpicture}
    \caption{Publications for NLP in general and references to popular Transformer papers.}
    \label{fig:scholar_scopus_papers}
\end{figure}
\vspace{-2.5\baselineskip}
\subsection{The NLP field Transformation} 
 Having a strong presence within the last decade, \textbf{Word embeddings} is a term where words that have the same meaning have a similar representation. In 2013, Mikolov introduced two different techniques for text vectorization: \textbf{Skip Gram} and \textbf{Common Bag Of Words} (CBOW) \cite{Mikolov_Sutskever_Chen_Corrado_Dean_2013}. Both of them were released in a library under a single name, ``\textbf{Word2Vec}'' and later in the same year some improvements for both of them were suggested in an attempt to remedy polysemy \cite{Mikolov_Chen_Corrado_Dean_2013}. Not long before, in 2011, the same author had also introduced a \textbf{Recurrent Neural Net Language Model}(RNNLM) being up to 15 times more efficient compared to past approaches \cite{Mikolov_Kombrink_Burget_Cernocky_Khudanpur_2011}. 
 
 In 2014 Pennington proposed an unsupervised learning algorithm for retrieving vector representations for words, named \textbf{GloVe} \cite{Pennington_Socher_Manning_2014}. \textbf{Recurrent/Recursive Networks and LSTMs}  are intriguing recent developments in ML with Sutskever suggesting the \textbf{Sequence to Sequence} (sec2sec) model\cite{Sutskever_Vinyals_Le_2014}. Also at the same year, Kalchbrenner proposed a \textbf{Dynamic Convolutional Neural Network}(DCNN) \cite{Kalchbrenner_Grefenstette_Blunsom_2014} and Kim explored a variety of classification tasks \cite{Kim_2014}.
 
  Dong  in 2015 introduced \textbf{multi-column convolutional neural networks} (MCCNNs) to analyze questions from multiple aspects and create their representations \cite{Dong_Wei_Zhou_Xu_2015}. Yin presented a comparative study between CNN and RNN for NLP summing up the progress up until then  \cite{yin2017comparative}. Upadhyay in 2017 introduced a new method for managing polysemy in word embeddings \cite{Upadhyay_Chang_Taddy_Kalai_Zou_2017}. Peters Suggested the Embeddings from Language Models in 2018 \cite{Peters_Neumann_Iyyer_Gardner_Clark_Lee_Zettlemoyer_2018}. The Same year Chen introduced a new \textbf{LSTM} that outperformed all previous models on Natural Language Inference. \cite{Chen_Zhu_Ling_Wei_Jiang_Inkpen_2017}.


\vspace{-2.5\baselineskip}

\subsubsection{Introducing Attention: the era of Transformers}\hspace*{\fill} \\

Bahdanau, based on the previously proposed Encoder-Decoder architecture, introduced the term \textbf{``Attention''}\cite{bahdanau-attention-2016}: an alignment score for each input word, based on the decoder's previous hidden state and the current input state of the sentence. Using this score, the decoder can decide which parts of the input sentence are the most important, without having to encode all of the input sentences into a fixed length vector.

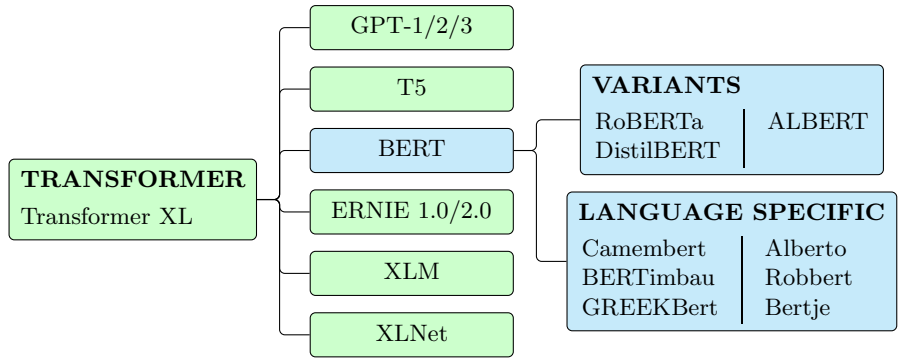
\begin{figure}[H]
    \begin{forest}
        forked edges,
        for tree={
            if level=0{fill=green!20}{
                if level=1{fill=green!20, minimum width=27mm}{
                    if level=2{fill=cyan!20}{}
                },
            },
            align=l,
            grow=east,
            inner sep = 1mm,
            parent anchor=east,
            child anchor=west,
            fork sep=3mm,
            draw,
            edge={rounded corners=2pt},
            rounded corners=2pt,
            node options={align=center,},
            l sep=7mm,
        }
        [\textbf{TRANSFORMER}\vspace{1mm}\\Transformer XL
            [XLNet]
            [XLM]
            [ERNIE 1.0/2.0]
            [BERT, fill=cyan!20
                [\textbf{LANGUAGE SPECIFIC}\vspace{1mm}\\
                    \begin{tabular}{l@{\hspace{3mm}}|@{\hspace{3mm}}l} 
                         Camembert & Alberto\\
                         BERTimbau & Robbert\\
                         GREEKBert & Bertje
                    \end{tabular}
                ]
                [\textbf{VARIANTS}\vspace{1mm}\\
                    \begin{tabular}{l@{\hspace{3mm}}|@{\hspace{3mm}}l} 
                    RoBERTa & ALBERT\\
                    DistilBERT
                    \end{tabular}
                ]
            ]
            [T5]
            [GPT-1/2/3]
        ]
    \end{forest}
    \caption{The evolution of Transformer based models.}
    \label{fig:transformers_bert_evolution}
\end{figure}

Following this concept, Vaswani\cite{vawsani-transformers-2019} made a rather bold proposal that leads to the \textbf{Transformers} architecture: the replacement of the costly RNNs with multi-headed self-attention layers in Encoder-Decoder models, thus increasing dramatically their performance, setting a new state-of-the-art for various tasks. Based on that idea, a whole new category of models emerged (Fig. \ref{fig:transformers_bert_evolution}). The \textbf{Transformer XL}\cite{dai-transformer-xl-2019} was suggested later on, attempting to resolve the limited length of the original Transformer's input.

\textbf{Transfer learning} and pre-training came along as another important progression. Howard and Ruder \cite{howard2018ulmfit} proposed Universal Language Model Fine-tuning (\textbf{ULMFiT}) in 2018, a transfer learning method that could be applied to any task in NLP, consisting of 3 stages: training in a large amount of text to capture general features, fine tuning for the task at hand with discriminative fine tuning and slanted triangular learning rates, and finally adding and fine tuning the classifier layers. Discriminative fine tuning allows the tuning of each layer with a different learning rate, while slanted triangular learning rates linearly increases the learning rate initially, linearly reducing it then again afterwards.

By combining the idea of pre-training and separating the Encoder part of this new Transformer architecture and stacking it as many times as needed, OpenAI's team (Radford \cite{radford-gpt-2018}\cite{radford-gpt-2019}, Brown \cite{brown-gpt-2020}) created 3 versions of the Generative Pre-Training model (a.k.a. \textbf{GPT-1/2/3}). Each version featured a larger number of parameters and pretraining in a larger corpus, achieving a new state-of-the-art for tasks like text generation and question answering with each version. The third version though is still not openly available, while a smaller model with 117 million parameters has been released for the second version.

\textbf{B}idirectional \textbf{E}ncoder \textbf{R}epresentations from \textbf{T}ransformers  (BERT) technique was introduced by Devlin\cite{devlin-bert-2019}, with an architecture similar to that of GPT. As the name suggests, one of its basic differences with GPT is the bi-directionality that helps in better understanding of the context, giving it a crucial advantage over other models. By releasing a base and an (extremely) larger model, BERT achieved state-of-the-art performance in tasks like question answering and text classification and can be used for a variety of other NLP tasks just by fine-tuning with a much smaller task-specific corpus.

Various publications branched from the initial BERT release, attempting to improve it or provide a solution to its drawbacks. \textbf{RoBERTa}\cite{liu-roberta-2019} was proposed as a better pretraining method, while \textbf{DistilBERT}\cite{sanh-distilbert-2019} and \textbf{ALBERT}\cite{lan-albert-2020} were smaller, faster alternatives with training speed and reduced memory consumption in mind.

Yang proposed \textbf{XLNet} \cite{yang2019xlnet}, an autoregressive (AR) model attempting to fix a discrepancy in the MLM task of BERT where the dependency between the masked input tokens is ignored. In order to achieve that, it’s using a permutation language modeling objective – meaning that all tokens are predicted instead of only the 15\% of BERT’s masked tokens. And though AR models can usually access the context in one direction, the permutation allows it to be bi-directional. XLNet can outperform BERT – sometimes significantly – at 20 tasks.

Sun from Baidu introduced Enhanced Representation through kNowledge IntEgration (\textbf{ERNIE}) \cite{sun2019erniev1} at the beginning of 2019, a character-based model antagonizing BERT for the current state-of-the-art with a slight different masking strategy - multi-stage instead of the random one BERT has. Later, in 2020, Baidu released a second version of ERNIE \cite{sun2019erniev2} introducing ``continual pretraining'' and multiple training tasks for lexical, syntactical and semantical analysis, claiming to outperform BERT and XLNet not only for the Chinese language, but for English as well (see Table \ref{table:glue_results}).

\vspace{-1.5\baselineskip}
\begin{table}[H]
\scriptsize
\def\arraystretch{1.5}
\setlength{\tabcolsep}{0.7mm}
\begin{tabularx}{\textwidth}{X|l|l|l|l|l|l|l|l|l|l} \toprule
& \bf{Score}& \bf{CoLA}   & \bf{SST-2}& \bf{MRPC}      & \bf{STS-B}     & \bf{MNLI-m}& \bf{QNLI}& \bf{RTE} & \bf{WNLI}& \bf{AX} \\
\rowcolor{black!5} ERNIE   
& \bf{90,9} & \bf{74,4}   & \bf{97,8} & \bf{93,9/91,8} & \bf{93,0/92,6} & \bf{91,9}  & \bf{97,3} & \bf{92,0} & \bf{95,9} & \bf{51,7} \\
\rowcolor{black!0} ALBERT
& -         &  69.1       & 97.1      & 93.4/91.2      & 92.5/92.0      & 91.3       & 91.0     & 89.2     & 89.2     & 50.2\\
\rowcolor{black!5} XLNet   
&   -       & 70.2        & 97.1      & 92.9/90.5      & 93.0/92.6      & 90.9       & -        & 88.5     & 92.5     & 48.4 \\ 
\rowcolor{black!0} RoBERTa 
& 88.1      & 67.8        & 96.7      & 92.3/89.8      & 92.2/91.9      & 90.8       & 95.4     & 88.2     & 89.0     & 48.7 \\
\rowcolor{black!5} XLM     
& 83.1      & 62.9        & 95.6      & 90.7/87.1      & 88.8/88.2      & 89.1       & 94.0     & 76.0     & 71.9     & 44.7 \\ 
\rowcolor{black!0} BERTlarge 
& 80.5      & 60.5        & 94.9      & 89.3/85.4      & 87.6/86.5      & 86.7       & 92.7     & 70.1     & 65.1     & 39.6 \\

\bottomrule
\end{tabularx}
\caption{Results for various Transformers in the current GLUE Leaderboard \cite{wang2018glue}.}\label{tab2}
\label{table:glue_results}
\end{table}

\vspace{-6.5\baselineskip}
\subsection{More Data, Please!}
\vspace{-2.5\baselineskip}
The General Language Understanding Evaluation \textbf{(GLUE)} benchmark is a  popular tool that evaluates the ability to analyze natural language understanding systems \cite{wang2018glue} featuring its very own leaderboard. A second version of it (\textbf{SuperGLUE}) came out a year later \cite{sarlin2020superglue}, featuring more and harder tasks, to achieve an even more accurate evaluation of the ever-evolving NLP models. GLUE consists of 11 tasks and their equivalent compilation of test datasets, while SuperGLUE features 10 more.

\textbf{Common Crawl} \cite{commoncrawl} is a repository with a significant amount of web crawl data, used as pre-training material for many of the models. It might be vast, but since it's a web dataset it's quite possible that heavy preprocessing is required before being actually usable. Other common sources of data between pretrained models are the Wikipedia pages - the main source for many multilingual and non-English models - as well as parsed subsets of the Reddit, IMDB, or Twitter websites. \textbf{Kaggle}\cite{kaggle2021} is another great source with a wide variety of user-submitted datasets, though due to their much smaller size they might be more suitable for fine tuning, rather than actually pretraining.

\vspace{-2.5\baselineskip}
\subsection{The Open Toolkits}
\vspace{-2\baselineskip}
In Table \ref{open_source_tools_table}, a summary of popular open source NLP tools is presented. Github repository stars are used as an indicator of popularity, while also making sure the projects are still active with recent/frequent releases.

Furthermore, \textbf{Huggingface} \cite{wolf-huggingface-2019} maintains a carefully curated git repository since 2019, with many of the latest pretrained Transformers for PyTorch and Tensorflow, allowing the quick testing of any of these models and turning prototyping into a breeze.

\begin{table}[H]
    \small
    \def\arraystretch{1.3}
    \setlength{\tabcolsep}{1mm}
    \begin{tabularx}{\textwidth}{lcX}
    \rowcolor{black!30}
                           & Stars & Description \\
    \rowcolor{black!0} \textbf{spaCy \cite{githubspacy}} & 19883 & Python library with pretrained models and great multilingual support, not suggested for research and benchmarking.\\
    \rowcolor{black!10} \textbf{Flair \cite{githubflair}} & 10091 & PyTorch library developed by Zalando, featuring the ``Flair embeddings'' for more efficient text vectorization.\\
    \rowcolor{black!0} \textbf{AllenNLP \cite{githuballen}} & 9794  & Designed for quick prototyping and research with a variety of pretrained models.\\
    \rowcolor{black!10} \textbf{NLTK \cite{githubNLTK}} & 9711 & One of the most recognizable libraries for NLP with some ML options - though not a common option for the task. \\
    \rowcolor{black!0} \textbf{Stanza \cite{githubstanza}} & 5271 & Stanford's Python library for ``Many Human Languages''. Can also be used as an interface to CoreNLP for even more features. \\
    \rowcolor{black!10} \textbf{SparkNLP \cite{githubsparknlp}} & 2002 & Built on top of Apache Spark and TensorFlow for speed and scalability, with generic and domain specific models available.\\
    
    \end{tabularx}
    \caption{An overview of popular open source NLP tools \& libraries.}
    \label{open_source_tools_table}
\end{table}

As one can easily see, there's a plethora of options - each with its own advantages and drawbacks. For instance, AllenNLP is more research/education oriented, while spaCy is probably a better choice for production, and NLTK might be harder to use. In the end, it all comes down to the application's requirements, and the developing team's personal preferences.
\vspace{-0.5\baselineskip}
\subsection{Challenges}

Although NLP has evolved a lot over the last years, there are still some challenges. All the unstructured context needs to be translated into meaningful defined data in order to perceive the intended meaning and entities, based on grammar and context.

Text-Mining is used to identify non-trivial patterns in text-data, starting with the Data Collection, by building a corpus, Data pre-Processing handles and manipulates the corpus using sub-processes of tokenization, normalization and substitution. Most of the raw data are not useful to define features, usually containing a lot of noise, so the ML model tends to become less effective and difficult to train. The initial goal is to go from chunks with text to a list of cleaned tokens, and then proceed to data exploration \& visualization, having a better dataset prior to building the model. Unstructured data transformed into useful text, by splitting the text into sentences, words and converted into standard form, like expanding constructions and set them to their base style.

The semantic meaning of words, ambiguity, grammar, or even slang is something that needs to be handled. In the next figure we are quoting some of the main challenges of NLP. 
\vspace{-0.5\baselineskip}
\subsubsection{Ambiguity}
exists at every level in linguistics, as shown above. In natural language, it is common, words to have multiple meanings according the context of the sentence (contextual words). Contrariwise, different words can have the same meaning(synonyms). Irony, sarcasm and humor may use words with a specific state, but in fact imply the opposite.  Some other main types of ambiguity are: Lexical ambiguity, where a single token can be presented as a verb, noun, or adjective, Semantic ambiguity refers to the conceptual situation described in a sentence having multiple interpretation\cite{bucaria2004lexical} and Syntactic ambiguity is happening when there is a double meaning in a sentence and the syntax principles of the language are not followed.  Additionally, errors, typos, slang and inconsistencies complicates the translation.\cite{macdonald1994lexical} Many times in a sentence there is a discrepancy between the actual meaning and what is written. Pragmatics is the study dealing with this case.\cite{leech2016principles}.

\begin{figure}[H]
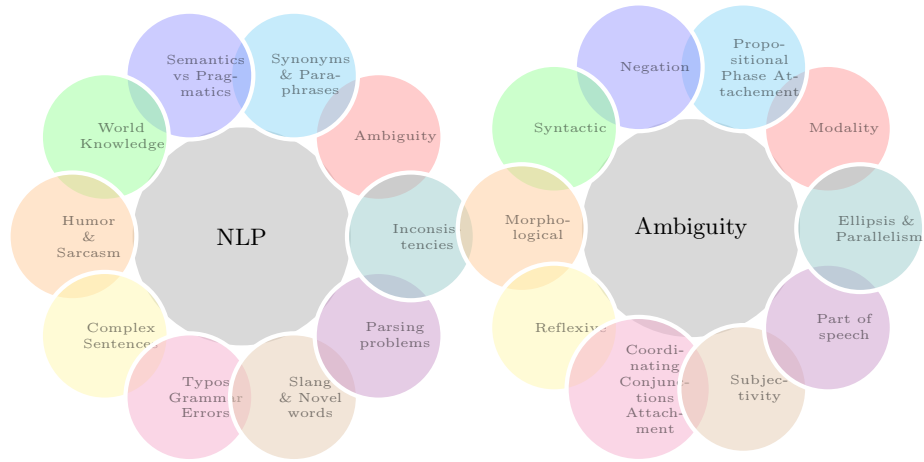


\smartdiagramset{
      bubble center node font = \small,
      bubble node font = \tiny,
      bubble center node size =2.9cm,
      bubble node size =1.7cm
    }
    \tikzset{
      bubble node/.append style={
        text width=1.1cm,
        align=center}
    }
\hskip-0.1cm
\smartdiagram[bubble diagram]{ NLP, Ambiguity,Synonyms\\ \& Paraphrases,Semantics \\vs Pragmatics, World \\Knowledge, Humor \\ \& Sarcasm, Complex\\ Sentences, Typos\\ Grammar \\Errors, Slang\\ \& Novel words, Parsing\\ problems, Inconsistencies}
\hskip-0.5cm
\smartdiagram[bubble diagram]{Ambiguity, Modality,Propositional Phase Attachement, Negation, Syntactic, Morphological, Reflexive,  Coordinating\, Conjunctions\, Attachment,Subjectivity, Part of \\ speech, Ellipsis \& \\ Parallelism}
  
  \caption{NLP Linguistic Challenges while parsing human languages \cite{bates-yates}}
\end{figure}

When input data is not text but speech, another ambiguity issue occurs. Phonetics and phonology refers to how tokens sound like. Phonetics deals with the vocal properties and perception, meaning how they are produced, and phonology deals with the expression in relation to each other in a language \cite{kurdi2016natural},\cite{mallamma2014semantical}.\\
\vspace{-1.5\baselineskip}
\subsubsection{Go Big (or maybe not?)}\hspace*{\fill}\\

\begin{figure}[H]
    \begin{tikzpicture}
        \begin{axis}[
            scaled ticks=false,
            axis y line=left,
            axis x line*=bottom,
            width = 1\linewidth,
            height = 0.35\linewidth,
            bar width=11mm,
            xtick style={yshift=-1ex, anchor=south},
            legend style={at={(-0.3,-0.3)}, anchor=north west},
            symbolic x coords={ELMo, GPT, BERT-Large, RoBERTa, XLM, GPT-2, MegatronLM, T5, TuringNLG},
            xtick = data]
            \addplot[ybar, nodes near coords, fill=black!10] coordinates{
                (ELMo, 94) (GPT, 110) (BERT-Large, 340) (RoBERTa, 355)
                (XLM, 655) (GPT-2, 1500) (MegatronLM, 8300) (T5, 11000) (TuringNLG, 17000)
            };
            \addplot[draw=blue,ultra thick,smooth] coordinates{
                (ELMo, 94) (GPT, 110) (BERT-Large, 340) (RoBERTa, 355)
                (XLM, 655) (GPT-2, 1500) (MegatronLM, 8300) (T5, 11000) (TuringNLG, 17000)
            };
            \label{transformers_total_parameters_plot}
        \end{axis}
    \end{tikzpicture}
    \caption{Number of parameters for models over time (in millions) \cite{sanh-distilbert-2019}.}
    \label{fig:transformers_total_parameters}
\end{figure}
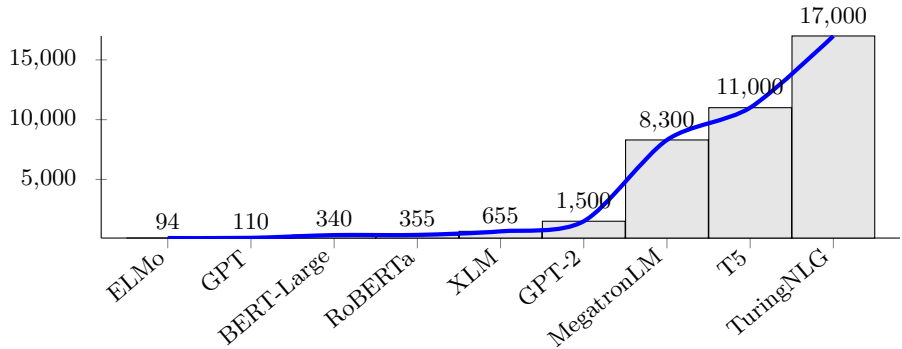

With every new publication, each team enters the race for a larger number of parameters (see Fig. \ref{fig:transformers_total_parameters}), ending up with models with billions of parameters. While that might improve the actual results, it has a huge impact on the training cost - both financially and time-wise - even if just fine tuning is required, making many of the latest architectures unusable for single GPUs or even whole GPU clusters in some cases. Besides whatever environmental consequences that might have, it's also \textbf{hindering their usage in real-world scenarios}.
\vspace{-1.5\baselineskip}
\vspace{-3\baselineskip}
\subsubsection{The Context Fragmentation \& Text Repetition}\hspace*{\fill} \\

The original Transformer architecture as well as popular deriving models has a predefined max input sequence length - in some cases defined by the model itself (e.g. 512 in BERT and GPT-1, 1024 for GPT-2) or by the overall hardware limitations (i.e. available memory). This might end up in \textbf{context loss in some marginal cases} where the input is split into segments where one's content correlates with its next or previous. Though in practice that's not usually a problem, there are some suggestions like the aforementioned Transformer XL that can help.

Furthermore, for the task of Question Answering (QA) or any other task that requires actual text generation, the output might be repetitive or even irrelevant on some occasions. Though sometimes that's due to the model itself or insufficient training, there are cases where the reason is this context fragmentation, or the lack of big/diverse enough datasets.
\vspace{-0.5\baselineskip}
\subsubsection{Lost in Translation}\hspace*{\fill} \\

Most of today's pretrained models are focusing on the English language, with some approaches focusing on character based languages like Chinese or Arabic. Some of the very few multilingual options are sparse pretraining attempts of BERT or GPT for local languages with monolingual models as a result (see Fig.\ref{fig:transformers_bert_evolution}). The multilingual version of BERT (or \textbf{m-BERT}), trained on the Wikipedia corpora, is another such option, supporting 104 different languages. The lack of a big enough, quality, multi-lingual dataset remains though, since even in Wikipedia many languages have a significantly lower number of pages (sometimes too low to even be considered) compared to English.

The \textbf{XLM model} \cite{conneau2020xlmr} attempts to take multilinguality one step further by adding byte pair encoding and training BERT in two languages at once, while changing the ``Masked Language Modeling'' objective to ``Translation Language Modeling'', masking tokens in either of the two languages. While that creates a multi-language model that can outperform m-BERT, it makes the data availability even harder, since now the same content is required in two languages at once. \textbf{XLM-R} \cite{lample2019xlm} is an XLM successor, using a much larger dataset and improving performance at a scale that can be fine tuned for one language and then be used for cross-lingual tasks.

While all of the above do wield interesting results, there's still a huge gap between English and other languages in some tasks like QA. For instance, a F1/EM (Exact match) score of 80.6/67.8 in English becomes 68.5/53.6 in German for the MLQA question answering benchmark \cite{conneau2020xlmr}.
\vspace{-1\baselineskip}
\section{Conclusion}
During the last decade there has been a tremendous progress in the field of NLP whether it's overall improvements, or task specific. It started with the vectorization revolution, with new suggestions for the most important task at the core of every NLP pipeline, ending up to the recent introduction of Transformers and transfer learning that marked the beginning of the ``Golden Age''.

Over this period, more and more companies started to adopt Chatbot technologies. Starting with simple pattern matching support agents, they're getting more and more sophisticated, gradually starting to adopt Machine Learning techniques and models, making them even more human-like and shifting to actual learning, away from patterns.

Machine Translation is also progressing steadily. Being one of the main tasks for the Encoder-Decoder model and their Transformer successors, automated translation applications are nowadays more accurate than ever. As the text generation models evolve, automated article generation applications start to emerge and virtual copywriters indistinguishable from humans, might soon be a reality.

But even though NLP has come this far, there are still many things to be done. Multilingual support models have still a long road ahead, and recent models will need heavy optimization and down-scaling or wider adoption of efficient hardware before they can be actually used in real-world applications. The Next Big Thing is still to be seen, and Transformers are currently leading the way.
\\\\
\large\textbf{Acknowledgements}\\
\normalsize
This work was supported by the MPhil program ``\textbf{Advanced Technologies in Informatics and Computers}'', hosted by the Department of Computer Science, International Hellenic University, Greece, Kavala.

%
%
\bibliographystyle{splncs04}
\bibliography{mybibliography}

\end{document}